\documentclass[lettersize,journal]{IEEEtran}
\usepackage{amsmath,amsfonts}
\usepackage{algorithm}
\usepackage{algpseudocode}
\usepackage{array}
\usepackage{textcomp}
\usepackage{stfloats}
\usepackage{url}
\usepackage{verbatim}
\usepackage{graphicx}
\usepackage{cite}
\usepackage{hyperref}
\usepackage{xcolor}
\usepackage{harpoon}
\usepackage{mathtools}
\usepackage{multirow}
\usepackage{enumitem}
\usepackage{float}
\usepackage{array}
\usepackage{makecell}
\usepackage{tabularx}

\usepackage[font=small]{caption}
\usepackage[font=scriptsize,labelformat=simple]{subcaption}

\makeatletter

\begin{document}

\title{Goal-Oriented Communications for Physical AI: Design and Testbed}

\author{Shutong Chen, Wenkai Zhang,  
Adnan Aijaz, Miao Guo,
and Yansha Deng
\thanks{
This work has received funding from the ARIA through the AI Scientist programme for the project “Self-Reflective AI scientist”, the UK EPSRC CHEDDAR Hub ref: EP/X040518/1 and EP/Y037421/1, the innovation programme under Grant No. 101292933—MAGIC-6G, the EPSRC ICASE Studentship and Toshiba Europe Ltd, and EPSRC under Grant EP/W004348/1. 

S. Chen, W. Zhang, M. Guo, and Y. Deng are with the Department of Engineering, King’s College London, London WC2R 2LS, U.K. (e-mail: shutong.chen@kcl.ac.uk; wenkai.2.zhang@kcl.ac.uk; 
miao.guo@kcl.ac.uk;
yansha.deng@kcl.ac.uk) (Corresponding author: Yansha Deng).

A. Aijaz is with the Bristol Research and Innovation Laboratory, Toshiba Europe Ltd., Bristol BS1 4ND,  U.K. (e-mail: adnan.aijaz@toshiba.eu).
}}


\addtolength{\topmargin}{-0.24in}
\addtolength{\textheight}{0.24in}
\IEEEaftertitletext{\vspace{-1.5\baselineskip}}
\maketitle

\begin{abstract}
Physical AI relies on frequently-updated, latency-sensitive video stream to perceive, reason, and interact with the physical world, resulting in strict latency requirements with much higher data volumes that existing 5G networks cannot support.
Goal-oriented communication (GoC) offers as a promising approach to solve this challenge by transmitting only task-relevant semantic representations. 
However, existing GoC frameworks were mainly evaluated in the simulations while their effectiveness has never  been validated in a practical deployment of  physical AI application.
In this work, we develop an end-to-end GoC testbed for Physical AI, which connects a PiPER robot arm equipped with an RGB-D camera and a 5G modem to an NVIDIA Jetson AGX Orin edge server through a 5G OpenAirInterface network. 
We propose and implement three GoC frameworks that transmit 3D bounding boxes, 2D scene graphs, and 3D scene graphs, as three types of semantic representations, respectively. They share the common functional modules designed for closed-loop Physical AI applications, including semantic extraction, full stack 5G transmission, language model inference, digital twin validation, and robotic control.
Extensive experiments on our testbed show that our GoC frameworks reduce the task completion time by up to 52.6\% and improve task success probability by up to 45\%, compared to the traditional framework that periodically transmits the raw image data. These results validate the practical effectiveness of our GoC framework and pave the way for efficient and reliable Physical AI applications over future 6G networks. Project website: \url{https://sites.google.com/view/goc-physical-ai-testbed}.
\end{abstract}


%

\section{Introduction}
Physical AI, defined as the integration of AI into physical machinery that enables autonomous robots to perceive, reason, and interact with the real world in real time\cite{nokia}, is a promising technology reshaping diverse industries, with its global market projected to approach \$178 billion by 2033 \cite{market}. While some Physical AI applications can perform perception and control entirely on-device, emerging architectures increasingly distribute sensing, AI inference, and decision-making across robots, edge platforms, and cloud servers, where communication becomes an integral part of the perception–decision–action loop.
This introduces a new traffic pattern that traditional best-effort delivery struggles to support, where high-volume visual data (e.g., videos and point clouds) must be transmitted at a high frequency as inputs to AI models under stringent latency requirements imposed by real-time decision-making.
One example is Vision Language Action (VLA)-driven robotic control, where maintaining a typical control frequency of 1-10 Hz requires Full-HD video transmission at 30 fps, resulting in a tremendous data rate of 1.44 Gbps \cite{11480097}.
Therefore, there is an urgent need to fundamentally redesign the  network for Physical AI beyond simply allocating more resource to deliver increasingly large volumes of data.

In this context, goal-oriented communication (GoC) is a promising paradigm by transmitting only task-relevant  semantic representation  to accomplish the underlying communication goal \cite{zhou2022task}.
Recent studies have shown that video content can be extracted as text descriptions \cite{model}, motion keypoints \cite{Jiang2023Keypoints}, and object layouts \cite{Park2025Visual} to support  control tasks in Physical AI.
More relevantly, our previous study \cite{FDR} developed a GoC framework that enables robots to autonomously detect and recover from unexpected execution failures.
By extracting and transmitting 3D scene graphs (3D-SG) as semantic representations and further integrating a Small Language Model (SLM) for robot motion generation,
we demonstrate through simulations that GoC substantially reduces communication and computation overhead while supporting reliable Physical AI operation.

However, despite the potential of GoC, its practical deployment for Physical AI remains  unexplored, with the following aspects requiring  real-world validation:
1) the end-to-end performance of Physical AI is determined by the highly-coupled Sensing-Communication-Computation-Control (S3C) loop, yet the contribution of each component remains experimentally unquantified. For example, transmitting abstracted semantic representations does not always contribute to faster task completion or higher task success probability due to additional semantic extraction latency; 
2) most existing works characterised communication latency using simplified channel models in simulations, while overlooking that real-world data transmission must consider  complete communication protocol stack and the impact of packet processing, buffering, and retransmission;
3) the practical semantic extraction and edge inference complexity and time have not been exploited in real world. It is unclear whether resource-constrained robotic onboard platforms can support real-time sensory data acquisition and semantic extraction, while the actual latency and complexity of computation modules, such as language model inference, may differ from analytical or simulation-based estimates due to specific hardware and runtime conditions.

To fill these gaps, we develop a general experimental testbed that connects a physical robot and a commercial edge server through a 5G OpenAirInterface (OAI) network for the real-world implementation and evaluation of GoC for Physical AI applications, which presents three main contributions:
\begin{itemize}
    \item Different from our previous study \cite{FDR} built on simulations, we implement an end-to-end GoC testbed for Physical AI. We implement onboard semantic extraction at a PiPER robot equipped with an RGB-D camera and a 5G modem, also language model inference, digital twin reconstruction, and trajectory replanning at the edge server equipped with an NVIDIA Jetson AGX Orin, and build 5G OAI network between Piper robot and edge server.
    \item We develop and implement GoC frameworks that transmit multiple semantic representations for Physical AI on our testbed, including 3D bounding box (3D-BBox), 2D scene graph (2D-SG), and 3D-SG, where we experimentally characterise their data sizes and full-stack communication latency, quantifying how they influence  semantic extraction, communication,  and edge inference.
    \item We conduct experiments over representative Physical AI tasks and present the latency breakdown across the complete S3C loop, together with the task success probability, showing that our GoC framework substantially reduces task completion time while improving the task success probability compared to the traditional S3C framework.

\end{itemize}

\section{System Overview}
In this section, we first introduce typical physical AI systems, then present the traditional S3C framework and our GoC framework for Physical AI, finally define performance metrics.

\subsection{Physical AI System}
We consider an autonomous wireless Physical AI system \cite{model} consisting of a \textbf{robot UE} and an \textbf{edge server}.
During task execution, the robot's onboard RGB-D camera  periodically captures an RGB image $\boldsymbol{R}_i  \in \mathbb{R}^{h\times w\times3}$ and a corresponding depth image $\boldsymbol{D}_ i \in \mathbb{R}^{h\times w}$ of the robotic workspace, where $i$ denotes the frame index, $h$ is the image height, and $w$ is the image width. 
To enable local robotic control, at the robot UE side, the RGB image is first used for object detection to identify the object set $\mathcal{O}_i=\{o_i^m\}_{m=1}^{N_i}$ in the robotic workspace, where $N_i$ denotes the detected object number, while the corresponding depth image is used to estimate their 3D positions $\boldsymbol{c}_i^m=[x_i^m,y_i^m,z_i^m]^{\mathrm T}$ through back-projection.
After local processing, the robot UE either transmits the RGB image directly to the edge server or further extracts and transmits task-relevant semantic representations.
Based on the received information, the edge server analyses the task state, determines the next robot action, and sends control commands back to the robot for execution. 
Due to environmental disturbances, perception errors, or execution deviations, unexpected faults (e.g., collisions) may occur and  interrupt normal task execution. In such cases, the received RGB image or semantic representation must further enable the edge server to detect the execution failure, generate an appropriate recovery action, allowing the robot to resume the interrupted task.



\subsection{Traditional S3C Framework for Physical AI}
The traditional S3C framework \cite{10802284}\cite{10990233} relies on periodic image transmission and Visual Language Model (VLM) inference to support Physical AI operation. 
Specifically, the robot UE compresses the captured RGB images into PNG format and  transmits them to the edge server periodically. 
Upon reception, the edge server feeds the RGB image together with the task instruction into the VLM to determine the next robot action. When an execution failure occurs, the VLM further detects the fault and generates the corresponding recovery motion. For example, if an apple is accidentally dropped during transportation, the generated recovery command  instructs the robot to regrasp the apple, e.g., \textit{robot:\{action: pick, target\_obj: apple\}}.
The resulting command is then transmitted back to the robot UE, which moves to the 3D position of the specified target object to execute the recovery action.
However, the frequent transmission of bandwidth-intensive RGB images and computationally-expensive VLM inference result in considerable communication and computation latency.

\vspace{-3ex}
\subsection{Semantic Representation Definition}
Instead of transmitting raw images, we define three alternative goal-oriented semantic representations (i.e., 3D-BBox, 2D-SG, and 3D-SG) for physical AI, as defined below.

\begin{itemize}
\item \textbf{3D Bounding Box:}
The 3D-BBox $\boldsymbol{b}_i^m=(l_i^m,\boldsymbol{c}_i^m,\mathcal{Q}_i^m)$ represents each detected object $o_i^m$ by its label $l_i^m$, 3D centre position $\boldsymbol{c}_i^m$, and an enclosing cuboid $\mathcal{Q}_i^m$ defined by its eight vertices in the 3D workspace.
Accordingly, the transmitted 3D-BBox set at frame $i$ is defined as
\begin{align}
    \mathcal{B}_i
    =
    \left\{
    \boldsymbol{b}_i^m
    \right\}_{m=1}^{N_i}.
\end{align}

\item \textbf{2D Scene Graph:}
The 2D-SG is formatted as an undirected graph composed of nodes and edges, where each node corresponds to a detected object and each edge describes a spatial relationship between two objects. For an arbitrary object pair $(o_i^m,o_i^n)$, their labels $l_i^m$ and $l_i^n$ together with the 2D spatial relationship $r_{i,\text{2D}}^{m,n}$ form a triplet $\boldsymbol{s}_{i,\text{2D}}^{m,n}=(l_i^m,r_{i,\text{2D}}^{m,n},l_i^n)$. The transmitted 2D-SG at frame $i$ is then defined as a set of relationship triplets
\begin{align}
    \mathcal{S}_i^{\mathrm{2D}}
=
\left\{
\boldsymbol{s}_{i,\text{2D}}^{m,n}
\right\}_{(o_i^m, o_i^n)\in\mathcal{O}_i}.
\end{align}

\item \textbf{3D Scene Graph:}
The 3D-SG also adopts the graph structure with each triplet defined as $\boldsymbol{s}_{i,\text{3D}}^{m,n}$. 
However, its spatial relationships  are inferred in 3D space reconstructed from point clouds, which extends image-plane relationships in 2D-SG (e.g., beside/overlapping) into depth-aware 3D  relationships (e.g., in front of/behind), providing richer geometric details for   Physical AI reasoning and control.
The 3D-SG at frame $i$ is given by
\begin{align}
    \mathcal{S}_i^{\mathrm{3D}}
=
\left\{
\boldsymbol{s}_{i,\text{3D}}^{m,n}
\right\}_{(o_i^m, o_i^n)\in\mathcal{O}_i}.
\end{align}
\end{itemize}

\begin{figure}
\centering
\includegraphics[width=1\linewidth]{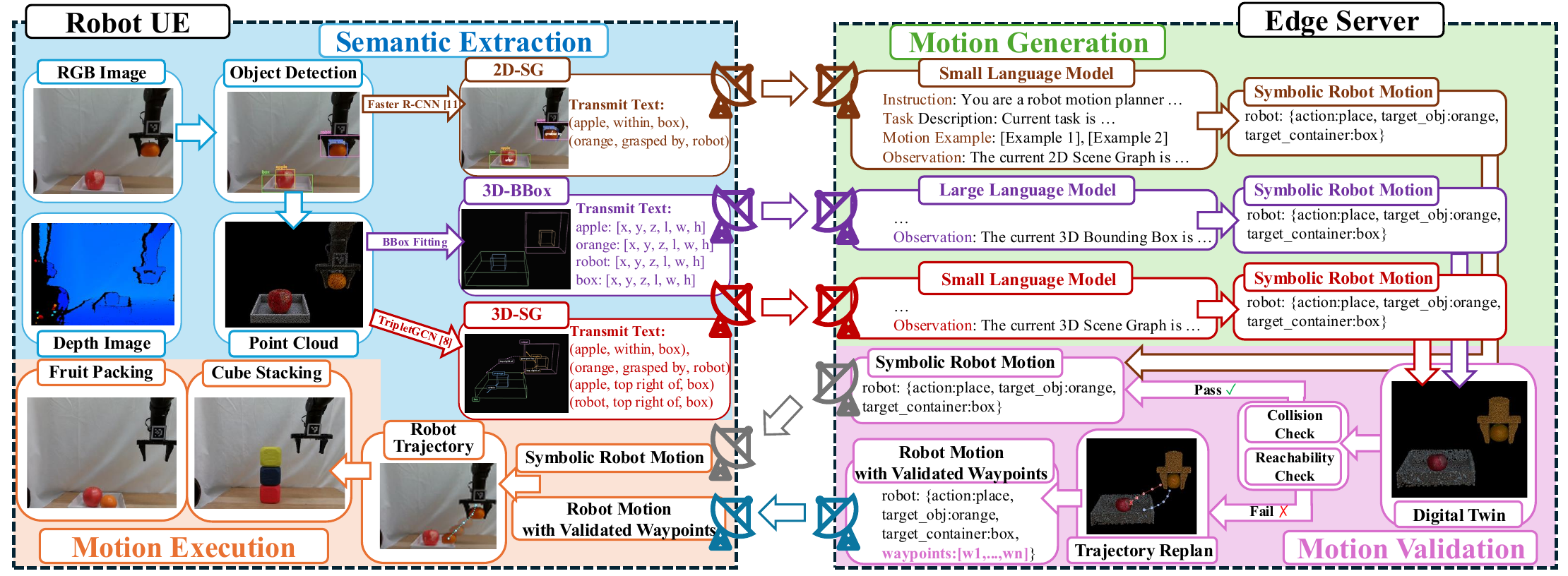}
\vspace{-3ex}
\caption{Our goal-oriented communication frameworks.}
\label{fig:system}
\vspace{-3ex}
\end{figure}

\vspace{-3ex}
\subsection{Goal-Oriented Communication Framework for Physical AI}
Based on these semantic representations, we develop GoC framework for Physical AI with a modular design: semantic extraction, uplink transmission, motion generation, motion validation, downlink transmission, and motion execution.

\textbf{Semantic Extraction:}
For 3D-BBox, based on the detected object set $\mathcal{O}_i$, the depth values within the detected region of each object $o_i^m$ are back-projected into the 3D workspace using the camera intrinsics, forming its corresponding 3D point cloud set. An enclosing cuboid $Q_i^m$ is then fitted to the resulting 3D points to represent the spatial extent of the object.

For 2D-SG, the masks of the detected object set $\mathcal{O}_i$ are processed using Faster R-CNN \cite{2DSG}, which extracts the image features of each object pair $(o_i^m,o_i^n)$ and their joint region to infer the 2D spatial relationship $r_{i,2D}^{m,n}$. The inferred relationships are combined with the corresponding object labels $l_i^m$ and $l_i^n$ to construct the 2D-SG representation $\mathcal{S}_i^{2D}$.

For 3D-SG, the same RGB-D preprocessing as that used for 3D-BBox is first applied to obtain the object point cloud sets. 
Therefore, it is important to note that the 3D-BBox $\mathcal{B}_i$ is inherently obtained during the extraction of the 3D-SG.
Then, for each object pair $(o_i^m,o_i^n)$, their individual point sets together with their merged point set are processed using TripletGCN \cite{FDR} to predict the corresponding 3D spatial relationship $r_{i,3D}^{m,n}$, which is then combined with the object labels $l_i^m$ and $l_i^n$ to construct the 3D-SG representation $\mathcal{S}_i^{3D}$.

\textbf{Uplink Transmission:}
After semantic extraction, the robot UE continuously transmits the latest semantic representation to the edge, ensuring that edge computation is always performed based on the most up-to-date data.
Specifically, the 3D-BBox-based GoC framework and 2D-SG-based GoC framework transmit 3D-BBox set $\mathcal{B}_i$ and 2D-SG set $\mathcal{S}_i^{2D}$, respectively, while the 3D-SG-based GoC framework transmits both 3D-SG set $\mathcal{S}_i^{3D}$ and its inherently available 3D-BBox set $\mathcal{B}_i$. 

\textbf{Robot Motion Generation:}
Upon receiving semantic representations, the edge server generates the symbolic robot motion using language model, such as  \textit{robot:\{action: pick, target\_obj: orange}\}.
For 3D-BBox, the received BBox is processed by the local Large Language Model (LLM) \footnote{Reasoning over 3D-BBox involves complex numerical interpretation that goes beyond the SLM's capability, therefore we use the more capable LLM.}, which interprets the object locations and spatial extents to generate the robot motion. 
For 2D-SG and 3D-SG, the relationship triplets $\mathcal{S}_i^{2D}$ and $\mathcal{S}_i^{3D}$ are processed by a fine-tuned SLM tailored for reasoning over symbolic scene relationships.

When execution failures occur, the semantic representation enables the language model to identify the abnormal task state. 
For 3D-BBox, if the position of an apple remains unchanged after a pick command, the LLM identifies the unsuccessful grasp and generates a new command to regrasp the apple. 
For 2D-SG and 3D-SG, the SLM identifies the failure from an unexpected relationship transition. For instance, after a pick command, the expected relationship \textit{$(\text{apple},\text{standing on},\text{table})$} should change to \textit{$(\text{apple},\text{grasped by},\text{robot})$}; otherwise, the SLM commands the robot to pick the apple again.

\textbf{Robot Motion Validation:}
For 3D-BBox and 3D-SG, the robot motion generated by language model  is further validated by a lightweight digital twin. Specifically, the edge server maintains a pre-collected library of 3D object models, from which the model of each detected object is selected according to its label, rescaled to fit the associated 3D-BBox $b_i^m$, and placed at detected position.
We also use the Unified Robot Description Format (URDF) model of the robot's end-effector (i.e., gripper) to reconstruct its geometry, with its pose updated based on its 3D-BBox.
The robot motion is then validated in digital twin by checking potential collisions between the gripper and objects. If no collision detected, the original motion is retained. Otherwise, the digital twin replans the motion with intermediate waypoints to avoid collisions, e.g., \textit{robot:\{action: pick, target\_obj: orange, waypoints: [$\mathbf{w}_1, ..., \mathbf{w}_n$]\}}.


\textbf{Downlink Transmission and Robot Motion Execution:}
Finally, the robot either maps the high-level symbolic motions into executable trajectories, or directly executes fine-grained waypoint sequences refined by the digital twin.

\vspace{-2ex}
\subsection{Evaluation Metrics}
We evaluate the traditional and our  GoC frameworks using the transmitted data size, task completion time, and task success probability, which characterise communication efficiency, time efficiency, and reliability of Physical AI, respectively.

\subsubsection{Transmitted Data Size}
It is measured by the amount of data transmitted from the robot UE to the edge server.

\subsubsection{Task Completion Time}
It is defined as the end-to-end elapsed time to complete a task.
For traditional S3C framework, it consists of uplink image transmission time $t_{\text{ul}}$, edge-side VLM inference time $t_{\text{VLM}}^{\text{edge}}$, downlink transmission time $t_{\text{dl}}$, and robot motion execution time $t_{\text{exe}}^{\text{UE}}$, given by
\begin{equation}
t_{\text{total}}^{\text{traditional}}
=
t_{\text{ul}}
+
t_{\text{VLM}}^{\text{edge}}
+
t_{\text{dl}}
+
t_{\text{exe}}^{\text{UE}}.
\end{equation}
For our GoC framework, it consists of the UE-side semantic extraction time $t_{\text{semantic}}^{\text{UE}}$, the uplink and downlink transmission time $t_{\text{ul}}$ and $t_{\text{dl}}$, the edge-side language model inference time $t_{\text{LM}}^{\text{edge}}$, optional DT reconstruction time $t_{\text{DT}}^{\text{edge}}$, and the UE-side motion execution time $t_{\text{exe}}^{\text{UE}}$, which can be expressed as
\begin{equation}
\begin{aligned}
t_{\text{total}}^{\text{GoC}}
={}t_{\text{semantic}}^{\text{UE}}
+t_{\text{ul}}
+t_{\text{LM}}^{\text{edge}}
+t_{\text{DT}}^{\text{edge}}
+t_{\text{dl}}
+t_{\text{exe}}^{\text{UE}}.
\end{aligned}
\end{equation}

\subsubsection{Task Success Probability}
It is measured by the percentage of trials in which the Physical AI successfully completes the task despite unexpected execution failures.

\vspace{-1ex}

\section{Testbed Setup}
\begin{figure}
\centering
\includegraphics[width=0.85\linewidth]{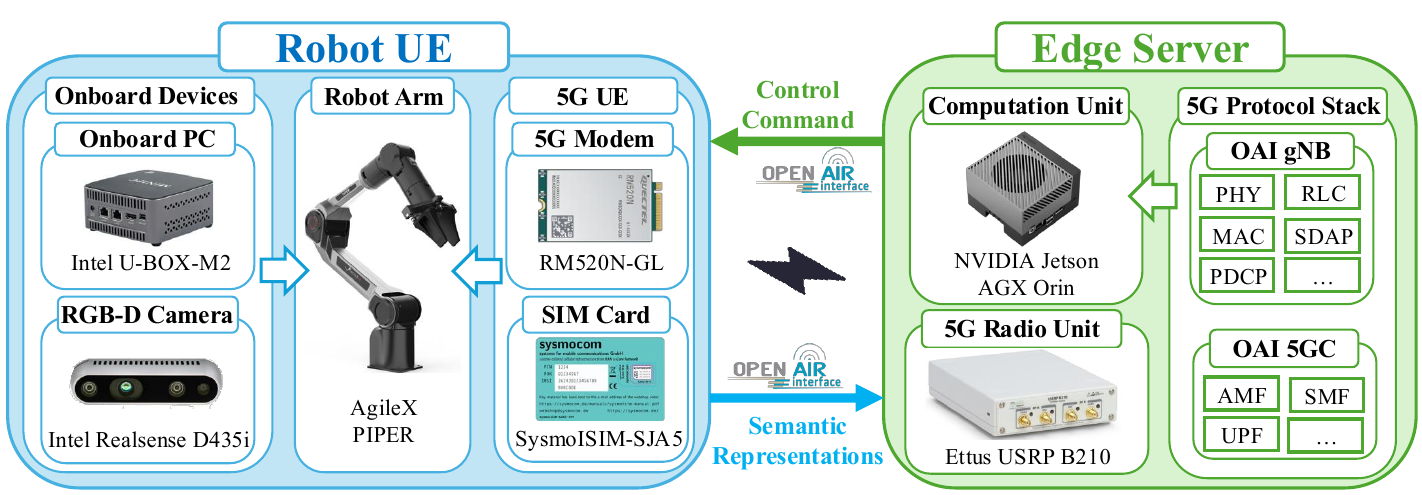}
\vspace{-0.5ex}
\caption{Testbed overview.}
\label{fig:testbed}
\vspace{-3ex}
\end{figure}

As shown in Fig. \ref{fig:testbed}, we present our experimental testbed for Physical AI, which consists of the robot UE, the edge server, and the bidirectional wireless communication between them. 

\vspace{-2ex}

\subsection{Robot UE Setup}
The robot UE consists of (i) an \textbf{Intel U-BOX-M2 mini PC} equipped with an Intel i7-8565U CPU, serving as the onboard computing platform,  (ii) an \textbf{AgileX PiPER robotic arm} with six degrees of freedom, equipped with a two-finger gripper, (iii) an \textbf{Intel RealSense D435i RGB-D camera} for acquiring colour and depth images, (iv) a \textbf{Quectel RM520N-GL 5G modem}, implementing the UE-side protocol stacks and radio functions, and (v) a \textbf{5G programmable Universal Subscriber Identity Module (USIM)} containing the subscriber identity.

\subsection{Edge Server Setup}
The edge server is deployed on a \textbf{NVIDIA Jetson AGX Orin} computing unit, connected to an \textbf{Ettus Universal Software Radio Peripheral (USRP) B210}, which integrates the functions of a 5G base station and an edge computing server.

\textbf{5G Base Station:} The Jetson AGX Orin, running the NVIDIA Sionna Research Kit \cite{Sionna}, deploys dockerised 5G core network (5GC) and  the 5G OAI gNB. 
The 5GC is implemented using the \textit{oai-amf}, \textit{oai-smf}, and \textit{oai-upf} containers to enable UE management, Protocol Data Unit (PDU) session management, and user-plane packet forwarding, while the gNB is instantiated in the \textit{oai-gnb} container and executes the CUDA-accelerated 5G  New Radio (NR) protocol stack.
The USRP B210 interfaces with the gNB through the USRP Hardware Driver (UHD), and serves as the Radio Unit (RU) for conversion between digital baseband and radio frequency signals and for over-the-air radio transmission and reception.

\textbf{Edge Server:} The Jetson AGX Orin is equipped with a 2048-core NVIDIA Ampere GPU and 64 GB of memory, and undertakes all edge computing tasks required for Physical AI, including 
(i) VLM and LLM inference using 4-bit quantised Qwen2.5-VL-72B model, 
(ii) SLM inference using 8-bit quantised Llama-3.2-1B model, 
(iii) digital twin reconstruction using MuJoCo simulator \cite{mujoco}, 
and (iv) collision-aware robot trajectory replanning within the digital twin using the rapidly-exploring random tree connect (RRT-Connect) algorithm \cite{RRT}.

\begin{table}[t]
\centering
\caption{5G NR Communication Setup.}
\label{tab:1}
\renewcommand{\arraystretch}{1}
\setlength{\tabcolsep}{2.5pt}
\small
\begin{tabularx}{\columnwidth}{
|>{\centering\arraybackslash}X
|c
|>{\centering\arraybackslash}X
|c|}
\hline

Bandwidth
& 40 MHz
& Distance
& 0.5 m \\ \hline

Resource Blocks
& 106
& Numerology
& 1 \\ \hline

Subcarrier Spacing
& 30 kHz
& Carrier Frequency
& 3.3 GHz \\ \hline

Downlink Band
& n77
& Uplink Band
& n77 \\ \hline

Transmit Antennas
& 2
& Receive Antennas
& 2 \\ \hline

\raisebox{-0.15ex}{TX/RX Gain}
& \raisebox{-0.15ex}{70/40 dB}
& \raisebox{-0.15ex}{EIRP}
& \raisebox{-0.15ex}{11.8–13.2 dB} \\ \hline

\end{tabularx}
\vspace{-0.3cm}
\end{table}

\subsection{5G Wireless Communication}
The bidirectional 5G NR standalone link is established between the Quectel RM520N-GL modem and the OAI gNB, with detailed radio configurations  summarised in Tab \ref{tab:1}. Before connection, we first add the subscriber profile corresponding to the programmable USIM to the OAI 5GC database, including the International Mobile Subscriber Identity (IMSI), authentication key, and Operator Variant Algorithm Configuration Field (OP).  
After network registration and authentication, the robot UE is assigned an IP address and establishes a full-duplex TCP connection with the edge server.

\section{Experiments}
This section evaluates the performance of the traditional S3C framework and our GoC framework on our testbed.
We consider two representative Physical AI tasks, each evaluated over 20 independent runs per framework: 1) \textbf{Fruit Packing}, requiring the robot to pick up an apple and an orange and place them inside a box with limited space.
2) \textbf{Cube Stacking}, requiring the robot to stack three coloured cubes in user-specified order.
We also introduce random faults during the task execution, including \textbf{Grasp Failure},  \textbf{Object Drop}, \textbf{Incorrect Placement}, \textbf{Incorrect Stacking Orders}, and \textbf{Collision}.


\begin{table*}[t]
\centering
\caption{Communication performance of different transmitted data over Wi-Fi and 5G NR links.}
\renewcommand{\arraystretch}{1.05}
\setlength{\tabcolsep}{1.5pt}
\label{tab:coms}
\scriptsize
\begin{tabular}{|c|c|c|c|c|c|c|c|c|c|c|c|c|c|}
\hline
\rule[-0.8ex]{0pt}{4.5ex}
\multirow{2}{*}{\shortstack{\textbf{Data}\\\textbf{Type}}} &
\multirow{2}{*}{\shortstack{\textbf{PNG}\\\textbf{Compression}\\\textbf{Level}}} &
\multirow{2}{*}{\shortstack{\textbf{Data}\\\textbf{Size}\\\textbf{(KB)}}} &
\multirow{2}{*}{\shortstack{\textbf{PNG}\\\textbf{Encoding}\\\textbf{(ms)}}} &
\multicolumn{3}{c|}{
\raisebox{0.5ex}{\textbf{TCP Buffering (ms)}}
} &
\multicolumn{3}{c|}{
\raisebox{0.5ex}{\textbf{Communication Time (ms)}}
} &
\multirow{2}{*}{\shortstack{\textbf{PNG}\\\textbf{Decoding}\\\textbf{(ms)}}} &
\multicolumn{3}{c|}{
\raisebox{0.5ex}{\textbf{Total Latency (ms)}}
}
\\
\cline{5-10}
\cline{12-14}
& & & &
\rule[-1.3ex]{0pt}{6ex}
\shortstack{\textbf{Wi-Fi}\\\textbf{(20 MHz)}} &
\shortstack{\textbf{5G}\\\textbf{(20 MHz)}} &
\shortstack{\textbf{5G}\\\textbf{(40 MHz)}} &
\shortstack{\textbf{Wi-Fi}\\\textbf{(20 MHz)}} &
\shortstack{\textbf{5G}\\\textbf{(20 MHz)}} &
\shortstack{\textbf{5G}\\\textbf{(40 MHz)}} &
&
\shortstack{\textbf{Wi-Fi}\\\textbf{(20 MHz)}} &
\shortstack{\textbf{5G}\\\textbf{(20 MHz)}} &
\shortstack{\textbf{5G}\\\textbf{(40 MHz)}}
\\
\hline
\multirow{3}{*}{\shortstack{RGB\\Image}}
& 0
& 901.97
& 32.39
& 165.62
& 174.91
& 140.37
& 764.27
& 812.55
& 268.41
& 3.26
& 965.54
& 1023.11
& 444.43
\\
\cline{2-14}
& 3
& 184.74
& 100.83
& 4.78
& 5.26
& 2.03
& 212.63
& 218.37
& 154.79
& 5.89
& 324.13
& 330.35
& 263.54
\\
\cline{2-14}
& 9
& 150.06
& 543.66
& 0.77
& 0.84
& 0.08
& 90.35
& 97.82
& 48.96
& 5.16
& 639.94
& 647.48
& 597.86
\\
\hline
3D-BBox
& N/A
& 0.7794
& N/A
& 0.13
& 0.12
& 0.07
& 21.10
& 20.42
& 19.32
& N/A
& 21.23
& 20.54
& 19.39
\\
\hline
2D-SG
& N/A
& 0.0383
& N/A
& 0.09
& 0.10
& 0.09
& 17.82
& 17.96
& 17.75
& N/A
& 17.91
& 18.06
& 17.84
\\
\hline
3D-SG
& N/A
& 0.0527
& N/A
& 0.10
& 0.10
& 0.09
& 20.47
& 21.58
& 18.08
& N/A
& 20.57
& 21.68
& 18.17
\\
\hline
3D-SG\&3D-BBox
& N/A
& 0.8823
& N/A
& 0.17
& 0.15
& 0.09
& 18.40
& 21.99
& 19.42
& N/A
& 18.57
& 22.14
& 19.51
\\
\hline
\end{tabular}
\end{table*}

\begin{figure*}[t]
    \centering

    \newcommand{\subcapshift}{3mm}   
    \newcommand{\subcapskip}{2pt}    
    
    \captionsetup[subfloat]{
    justification=centering,
    singlelinecheck=false,
    labelfont=bf,
    textfont=bf,
    margin={\subcapshift,0mm},
    skip=\subcapskip
    }

    \subfloat[Semantic extraction time.]{
        \includegraphics[width=0.235\textwidth]{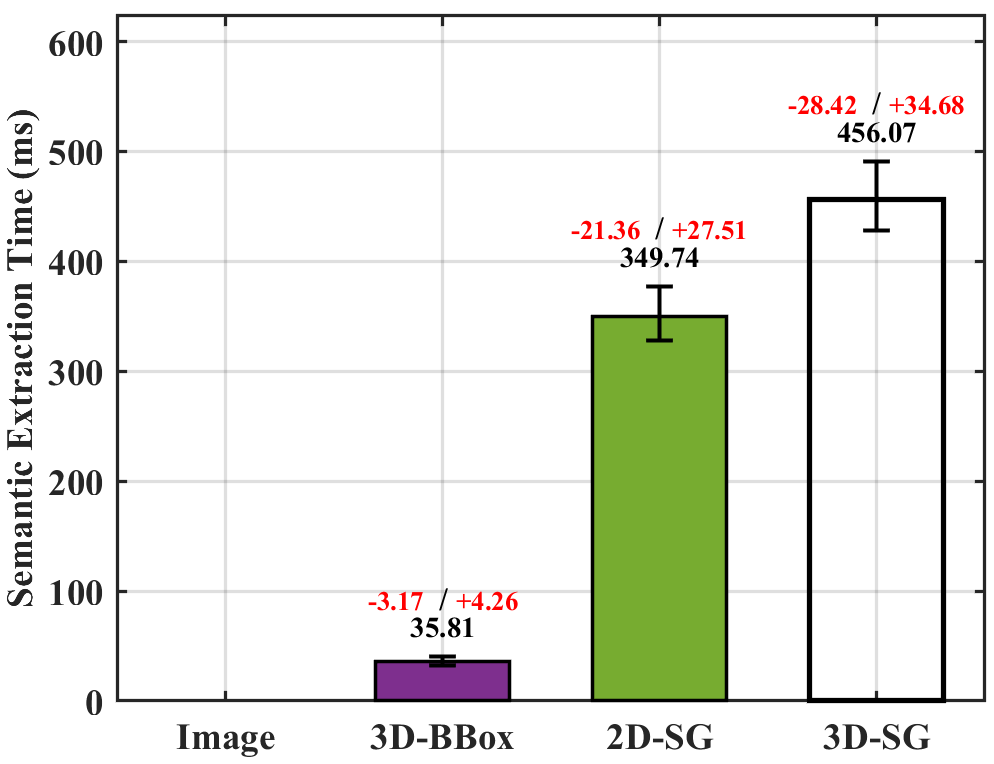}
    }
    \hfill
    \subfloat[Uplink transmission time.]{
        \includegraphics[width=0.235\textwidth]{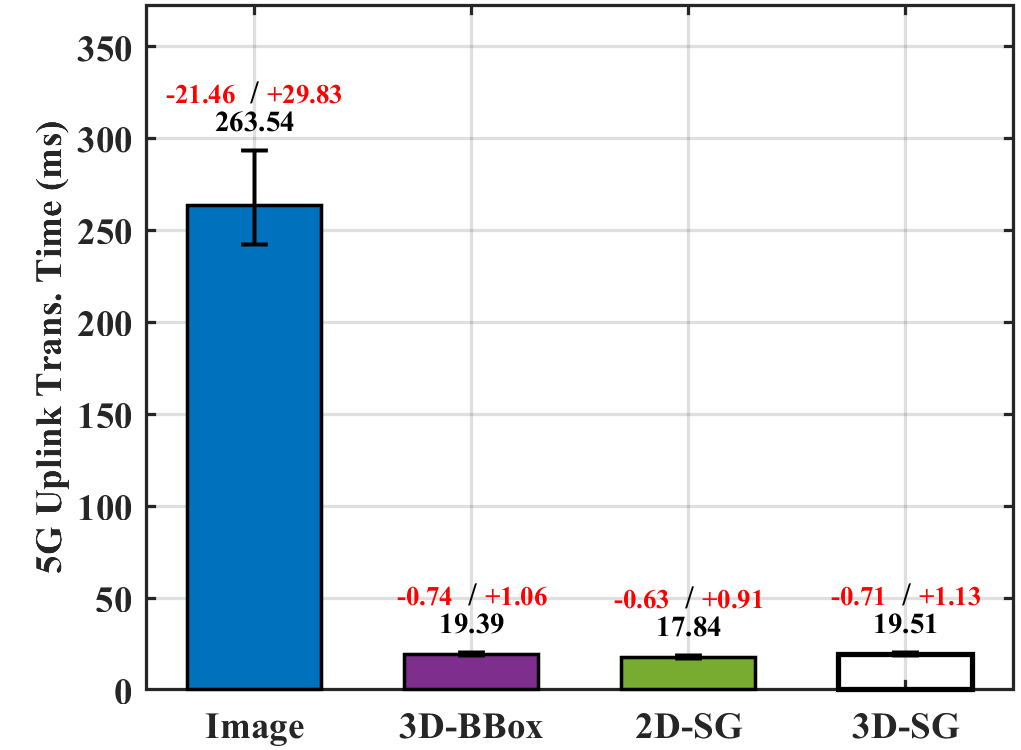}
    }
    \hfill
    \subfloat[Language model inference time.]{
        \includegraphics[width=0.235\textwidth]{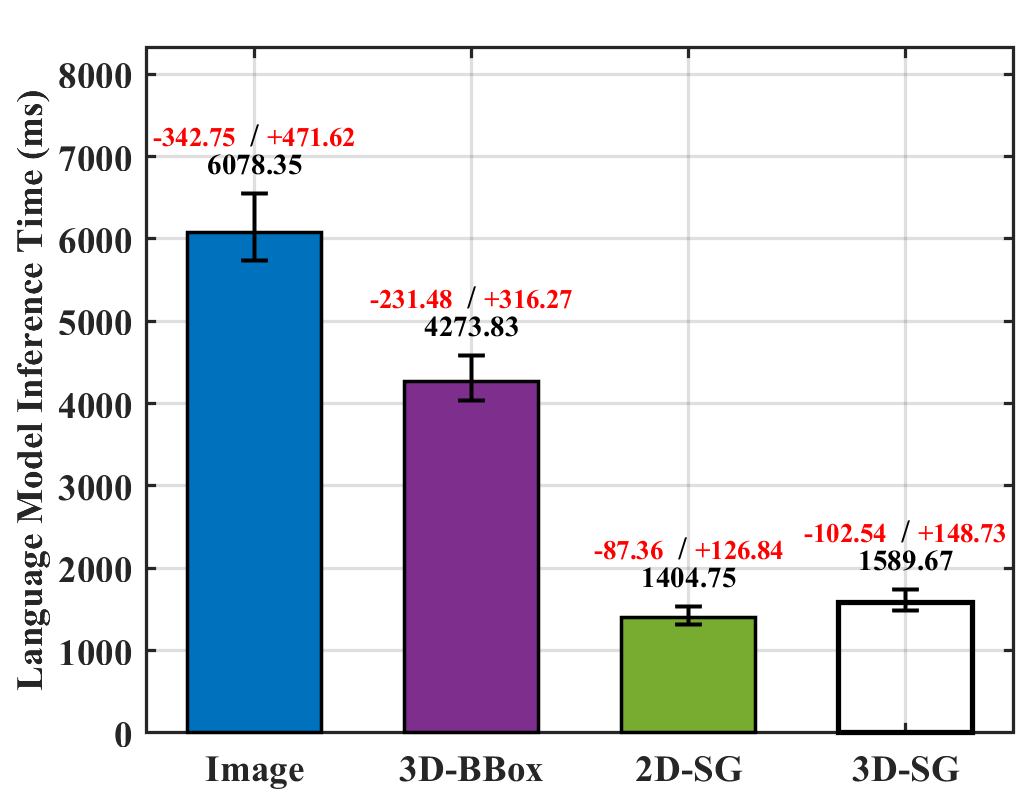}
    }
    \hfill
    \subfloat[Digital twin reconstruction time.]{
        \includegraphics[width=0.235\textwidth]{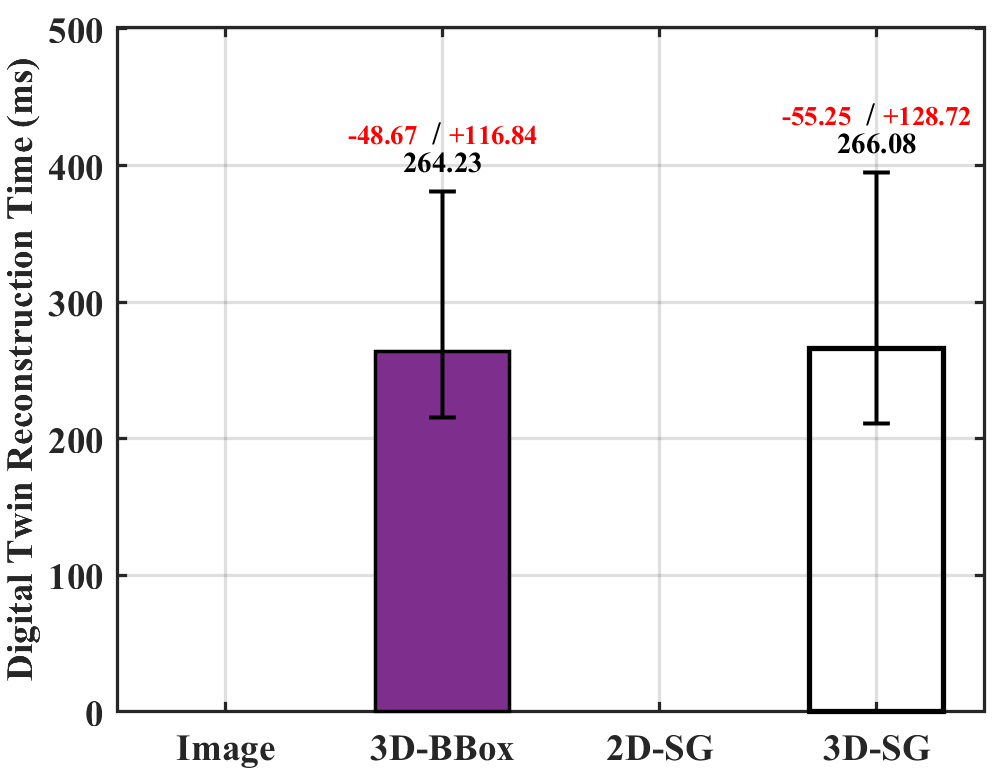}
    }

    \vspace{1mm}

    \subfloat[Downlink transmission time.]{
        \includegraphics[width=0.225\textwidth]{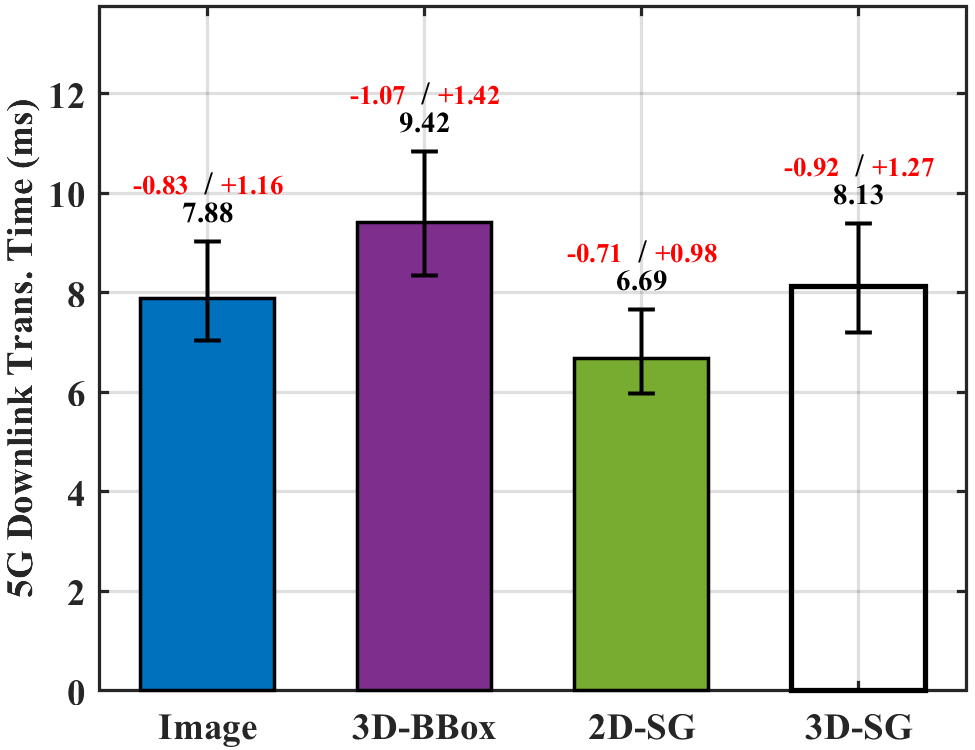}
    }
    \hfill
    \subfloat[Control time.]{
        \includegraphics[width=0.235\textwidth]{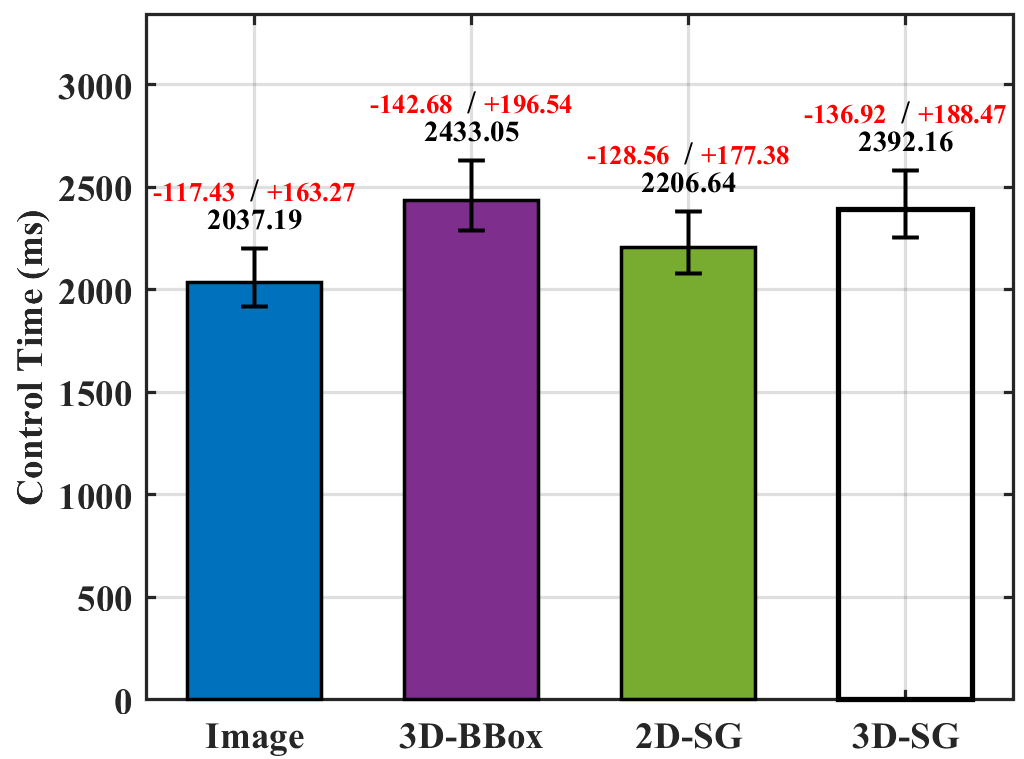}
    }
    \hfill
    \subfloat[Task completion time.]{
        \includegraphics[width=0.235\textwidth]{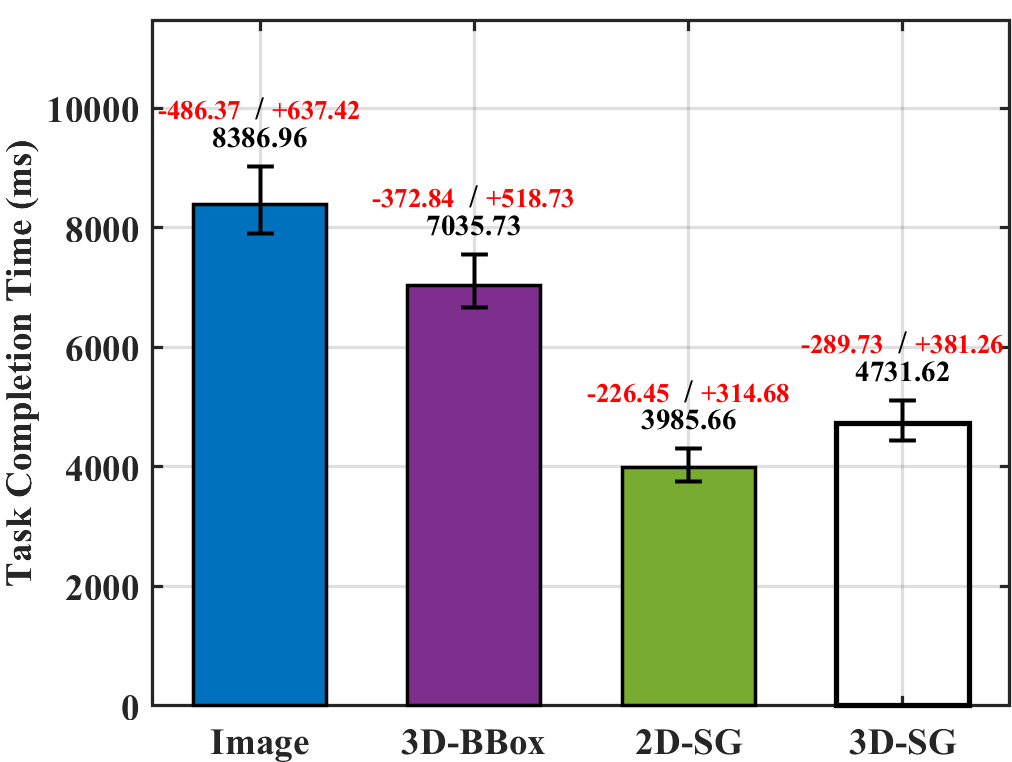}
    }
    \hfill
    \subfloat[Task success probability.]{
        \includegraphics[width=0.235\textwidth]{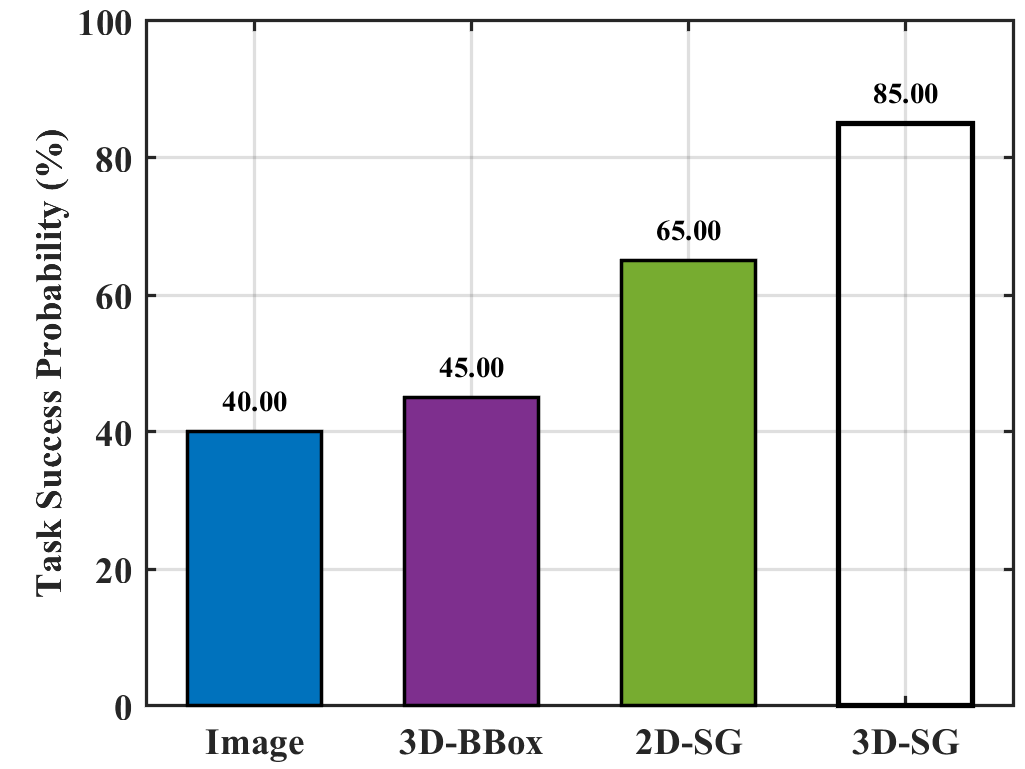}
    }

    \caption{Performance comparison of the traditional S3C framework and our proposed GoC framework.}
    \label{fig:sim}
    \vspace{-4mm}
\end{figure*}

Tab \ref{tab:coms} compares the communication performance of different data over Wi-Fi and 5G NR links under different bandwidth settings, where the Wi-Fi link is established using a GL-X3000 Wi-Fi 6 router, while the 5G NR link is additionally configured with 20 MHz bandwidth and 51 resource blocks.
It can be observed that: 1) compared with the RGB image at the highest PNG compression level, the data sizes of our semantic representations, 3D-BBox, 2D-SG, and 3D-SG, are reduced by 99.48\%, 99.97\%, and 99.96\%, respectively.
Consequently, their total communication latency is reduced by at least 96.65\% under both Wi-Fi and 5G NR links, which indicates the communication timeliness improvement of our GoC framework;
2) although increasing PNG compression level reduces the RGB image size and transmission time, it introduces considerable encoding latency on the resource-constrained robot UE. In contrast, a low compression level results in a significantly larger data size and therefore increases the TCP buffering and transmission latencies.
This proves that aggressively minimising the data size is not always optimal for communication efficiency due to the preprocessing overhead on robot UEs;
3) increasing the bandwidth from 20 to 40 MHz provides only a limited latency reduction for the compact semantic representations, as their communication time is already close to its practical lower bound, and the overall transmission latency is primarily determined by fixed protocol stack overhead, which again highlights the importance of accounting for full-stack communication overhead in practice.


Fig. \ref{fig:sim} presents the latency breakdown and task success probability of the traditional S3C framework and our three proposed  GoC frameworks. 
We can see from Fig. \ref{fig:sim}(a)-(c) that, although our GoC frameworks introduce additional semantic extraction latency, the compact semantic representations not only reduce the uplink transmission latency, but also reduce the language model inference latency at the edge. This is because the structured semantic representations directly encode explicit and interpretable spatial information required for Physical AI operation, thereby avoiding computationally-intensive VLM inference.
We can also see from Fig. \ref{fig:sim}(g)-(h) that our GoC frameworks reduce the task completion time by up to 52.6\% while improving the task success probability by up to 45\% compared to the traditional S3C framework. The reason is that the traditional framework experiences long communication and computation times that limit its timely response to time-sensitive faults such as collisions. On the other hand, raw RGB images contain excessive task-irrelevant visual information, increasing the perception and reasoning complexity while reducing the generated motion reliability. 
The 3D-BBox alleviates these issues by replacing VLM with LLM and introducing a digital twin to verify generated actions, but it still incurs relatively high inference latency, while the encoded object positions are insufficiently informative for control (e.g., it is difficult to tell whether  an object is grasped or not).
In comparison, 2D-SG explicitly encodes pairwise spatial relationships between objects that directly indicate the current task state, while maintaining low communication and computation overhead.
Further, 3D-SG extends these relationships into the more accurate depth-aware spatial information and, together with 3D-BBox, enables digital twin-based validation to improve the reliability of generated robot actions.

\section{Conclusion}
In this letter, we built an end-to-end Goal-oriented Communication (GoC) testbed for Physical AI, which closes the Sensing-Communication-Computation-Control loop between a PiPER robot arm and an NVIDIA Jetson AGX Orin edge server using 5G OpenAirInterface network. 
We designed and implemented three GoC frameworks on our testbed to transmit 3D bounding boxes, 2D scene graphs, and 3D scene graphs, respectively, together with the corresponding semantic extraction, 5G transmission, language model inference, digital twin validation, and robotic control modules. 
Experiments on representative Physical AI tasks showed that 
1) GoC offers a promising communication paradigm for Physical AI because it jointly designs communication with upstream sensing and downstream computation and control. That is, it transmits compact semantic representations that reduce data volume, and also preserves only  task-relevant information for efficient reasoning and reliable control;
2) in practice, reducing the transmitted data size does not always reduce end-to-end communication latency because data encoding and traversal through the full 5G protocol stack can introduce additional delays;
3) current best-effort video delivery cannot efficiently support Physical AI applications, not only because video traffic challenges existing networks, but also because processing raw video imposes considerable computational overhead that compromises information freshness for real-time control.

\bibliographystyle{IEEEtran}
\bibliography{IEEEabrv,testbed}

\end{document}